\documentclass[conference,letterpaper]{IEEEtran}

\usepackage[utf8]{inputenc} 
\usepackage[T1]{fontenc}
\usepackage{url}
\usepackage{ifthen}
\usepackage{cite}
\usepackage[cmex10]{amsmath} 

\usepackage{graphicx}

\def\pNMLSingle{p_{\hat{\theta}(z^N,x,y)} (y|x)}

\begin{document}
\title{A New Look at an Old Problem: A Universal Learning Approach to Linear Regression} 

\author{%
  \IEEEauthorblockN{Koby Bibas}
  \IEEEauthorblockA{School of Electrical Engineering\\
                    Tel Aviv University\\
                    Email: kobybibas@gmail.com}
  \and
  \IEEEauthorblockN{Yaniv Fogel}
  \IEEEauthorblockA{School of Electrical Engineering\\
                    Tel Aviv University\\ 
                    Email: Yaniv.fogel8@gmail.com}
  \and
  \IEEEauthorblockN{Meir Feder}
  \IEEEauthorblockA{School of Electrical Engineering\\
                    Tel Aviv University\\ 
                    Email: meir@eng.tau.ac.il}
}

\maketitle

\begin{abstract}
Linear regression is a classical paradigm in statistics. 
A new look at it is provided via the lens of universal learning.
In applying universal learning to linear regression the hypotheses class represents the label $y\in {\cal R}$ as a linear combination of the feature vector $x^T\theta$ where $x\in {\cal R}^M$, within a Gaussian error.
The Predictive Normalized Maximum Likelihood (pNML) solution for universal learning of individual data can be expressed analytically in this case, as well as its associated learnability measure. 
Interestingly, the situation where the number of parameters $M$ may even be larger than the number of training samples $N$ can be examined. 
As expected, in this case learnability cannot be attained in every situation; nevertheless, if the test vector resides mostly in a subspace spanned by the eigenvectors associated with the large eigenvalues of the empirical correlation matrix of the training data, linear regression can generalize despite the fact that it uses an ``over-parametrized'' model. 
We demonstrate the results with a simulation of fitting a polynomial to data with a possibly large polynomial degree.
\end{abstract}


\section{Introduction} \label{Introduction}
Linear regression, using least squares, is probably one of the most standard techniques in statistics, \cite{lawson1995solving}. This work provides a new view of this problem based on recent results in universal learning. In particular, the common assumption in linear regression is that the number of training samples needs to be higher than the number of features in order to be able to generalize \cite{james2013introduction}.
Recently, the success of Deep Neural Networks (DNNs) in which the number of learnable parameters may be greater by several orders of magnitudes than the size of the feature space, requires rethinking that assumption. 
The new view we provide will show that sometimes generalization can be attained even in the ``over-parameterized'' regime.

Before diving into this analysis, a short introduction to universal learning is provided. In the common situation of supervised machine learning, a training set is given consisting of $N$ pairs $z^N=\{(x_i, y_i)\}_{i=1}^{N}$, where $x \in {\cal X}$ is the data or the features and $y \in {\cal Y}$ is the label. Then, a new $x$ is given and the task is to predict its corresponding label $y$. 
In the information theoretic framework considered in a variety of works, e.g., \cite{universal_prediction} and more recently \cite{FogelFeder2018}, prediction is done by assigning a probability distribution $q(\cdot|x)$ to the unknown label, and the prediction loss is the log-loss:
\begin{equation}
\mathcal{L}(q;x,y) = -\log {q(y|x}).
\end{equation}
Clearly a reasonable goal is to find the predictor $q$ with the minimal loss for the test sample. However, this problem is ill-posed unless additional assumptions are made.

First, a ``model'' class, or `hypotheses'' class must be defined. 
This class is a set of conditional probability distributions
\begin{align} P_\Theta = \{ p_\theta(y|x),\;\;\theta\in\Theta\} \end{align} 
where $\Theta$ is a general index set. 
This is equivalent to saying that there is a set of stochastic functions  $\{ y=g_\theta(x),\;\;\theta\in\Theta\}$ used to explain the relation between $x$ and $y$.

Next, assumptions must be made on how the features and the labels are generated. 
In the stochastic setting, it is assumed that there is a true probabilistic relation between $x$ and $y$ given by an (unknown) model from the class $P_\Theta$. 
A more general setting, used in the variety of works in machine learning, is the Probably Approximately Correct (PAC) established in  \cite{valiant1984theory}. 
In PAC $x$ and $y$ are assumed to be generated by some source $P(x,y)=P(x)P(y|x)$, but unlike the standard stochastic setting $P(y|x)$ is not necessarily a member of the hypothesis class. 
In both the stochastic and PAC settings the goal is to perform as well as a learner that knows the true probability.

The most general setting, however, and the one used in this paper is 
the individual setting, where the features and the labels of both the training and test are specific, individual values.
In this setting the goal can no longer be to perform as well as a learner that knows the true probability. Instead, following \cite{universal_prediction}, the goal is to seek a learner that can compete with a ``genie'' or a reference learner that knows the desired label value, but is restricted to use a model from the given hypotheses class $P_\Theta$. In addition, as discussed in \cite{FogelFeder2018}, the reference does not know which of the samples is the test. Thus, the reference chooses
\begin{align} \hat{\theta}(z^N,x,y)  = \arg\max_\theta \left[ p_\theta(y|x) \cdot\Pi_{i=1}^N p_\theta(y_i|x_i) \right] \end{align}
The log-loss difference between a universal learner $q$ and the reference is the regret:
\begin{equation} \label{eq:genie_regret}
R(z^N,x,y,q) = \log \frac{p_{\hat{\theta}(z^N,x,y)}(y|x)}{q_(y|x;z^N)}.
\end{equation}
As advocated in \cite{FogelFeder2018}, the chosen universal learner solves:
\begin{equation} \label{eq:minmax_prob}
\min_q \max_y R(z^N,x,y,q) = R^*(z^N,x)
\end{equation}
Following \cite{shtar1987universal} this learner, called the Predictive Normalized Maximum Likelihood (pNML), is given by
\begin{equation} \label{eq:pNML}
q_{\mbox{\tiny{pNML}}}(y|x;z^N)=\frac{\pNMLSingle}{\sum_{y\in {\cal Y}} \pNMLSingle}.
\end{equation}
Note that this pNML probability assignment was essentially proposed earlier, see \cite{roos2008sequentially,roos2008bayesian}, with a different motivation as one of the possible variations of the Normalized Maximum Likelihood (NML) method of \cite{shtar1987universal} for universal prediction.

In order to obtain the pNML the following procedure is executed: assuming the label of the test data is known, find the best model that fits it with the training samples, and predict the assumed label by this model. 
Repeat the process for all possible labels. 
Then, normalize to get a valid probability distribution which is the pNML learner.
The regret of the pNML, $R^*(z^N,x)$ is the logarithm of its normalization factor
\begin{equation} \label{eq:pNML_regret}
R^*(z^N,x) = \log \left\{ \sum_{y\in {\cal Y}} \pNMLSingle \right\}.
\end{equation}

In considering linear regression, $y\in {\cal R}$ is the scalar label, $x\in {\cal R}^M$ is the feature vector (sometimes the first component of $x$ is set to $1$ to formulate affine linear relation), and the model class is the set:
\begin{equation} \label{regression_model}
\left\{ p_{\theta}(y|x) 
=\frac{1}{\sqrt[]{2\pi\sigma^2}}\exp\left\{-\frac{1}{2\sigma^2}\big(y- x^T\theta \big)^2\right\}, \;\; \theta \in {\cal R}^M \right\} 
\end{equation}
That is, the label $y$ is a linear combination of the components of $x$, within a Gaussian noise. As shown below, in this case the pNML and its learnability measure can be evaluated explicitly.

The pNML approach deviates from the standard Empirical Risk Minimization (ERM) \cite{vapnik1992principles} approach.
In ERM, given a training set and hypothesis class $\{p_\theta(y|x),\ \theta \in \Theta\}$, a learner that minimizes the loss over the training set is chosen:
\begin{equation}
q_{\mbox{\tiny{ERM}}}(y|x) = \underset{p_\theta}{\textit{argmin }} \frac{1}{N}\sum_{i=1}^{N}  \mathcal{L}(p_\theta; x_i, y_i).
\end{equation}
In the linear regression model (\ref{regression_model}), one chooses the least squares solution over the training set for the linear coefficients. 
This, however, may lead to large log-loss generalization error.

The paper has two main contributions.
First, it provides an explicit analytical solution for the pNML learner and its ``learnability'' measure (which is the minmax regret (\ref{eq:pNML_regret})) for the linear regression hypothesis class. This includes also the regularized case where the norm of the coefficients vector is constrained. Second, based on the analysis of the learnability measure, it is shown that even in the over-parameterized case where the number of parameters $M$ may be larger than the training size $N$, if the test data comes from a ``learnable space'' successful generalization occurs.
This phenomenon may explain why other over-parameterized models such as deep neural networks are successful for ``learnable'' data.

The paper outline is as follows.
Section \ref{sec:related_works} presents some related work. 
Section \ref{sec:formal_problem_def} provides the formal problem definition, while the pNML evaluation for the regression problem is given in sections \ref{sec:pNML_eval} and \ref{sec:pNMLwithReg}. 
In-depth analysis of the learnable space is given in section \ref{sec:learnable_space}. 
Simulation of the pNML and its regret for the problem of fitting a polynomial to data is described in section \ref{sec:simulation} and the conclusions are in section \ref{sec:conclusion}.

\section{Related Works} \label{sec:related_works}
In this section, we briefly mention related works on model generalization, least squares regression and the confidence of the least squares predictions.

\textbf{Model Generalization.} 
Understating the model generalization capabilities is considered a fundamental problem in machine learning  \cite{vapnik2013nature}. 
As noted, most of the theoretical work in learning use the PAC setting. In that setting, a common measure is the VC Dimension that can be used to upper bound on the test generalization error.
For DNNs, the VC dimension is linear with the number of parameters \cite{sontag1998vc}, yet the empirical evidence demonstrates that DNNs have state of the art generalization performance. This makes the VC dimension irrelevant for assessing the generalization error of DNNs.

\textbf{Least Squares.}
The least squares algorithm is widely used in linear regression due to its robust performance and simplicity of implementation.
In addition to the explicit formula for its solution, it can be solved sequentially, via the Recursive Least Squares (RLS) algorithm, which is an efficient online method for finding the linear predictor that minimizes the squared error over the training data \cite{hayes19969}.
This paper provides a new look at the classical least squares method, the individual setting using the pNML approach.

\textbf{Outliers Detection and Confidence.}
In order to evaluate a pointwise confidence measure for linear regression, several methods were proposed. 
Leverage values are employed to identify outliers with respect to their feature values \cite{cardinali2013observation}. 
A leverage value is a measure of the distance between an observation and the center of the data\footnote{Whenever a matrix is inverted it is assumed that the matrix is invertible. If needed, $\lambda I$ with small $\lambda$ is added to assure invertibility}
\begin{equation}
h_{ii}=x_i^T(XX^T)^{-1}x_i
\end{equation}
where $XX^T$ is the (unnormalized) correlation matrix of the training set and $x_i$ is the feature value which is examined.
If the leverage value $h_{ii}$ of observation is large, the observation is considered as an outlier.
Using the pNML, in section \ref{sec:pNML_eval} we get a confidence measure for the prediction of the next label which is similar to the leverage measure.

Another approach for finding the reliability of the prediction is to compose confidence intervals \cite{trevor2009elements}.
Confidence intervals are a pointwise measure that is sensitive to the variability of the features and sample size.
Denote $\hat{y}$ as the predicted value of $x$ and  $\hat{\sigma}^2$ as the empirical error of the prediction, under the assumption of stochastic i.i.d data and existence of white noise, a confidence interval convergences in distribution to
\begin{equation}
(\hat{y} - y) \xrightarrow{} \mathcal{N}(0, \hat{\sigma}^2 x^T(XX^T)^{-1}x).
\end{equation}

\section{Linear Regression: Formal Problem Definition} \label{sec:formal_problem_def}
Given N pairs of data and labels $\{x_i, y_i\}_{i=1}^{N}$ where $x_i \in R^M, y_i \in R$ are deterministic, the model takes the form:
\begin{equation}
\begin{split}
y_1&=x_1^T \theta + e_1 \\
   &\ \vdots \\
y_{N}&=x_{N}^T \theta + e_{N} \\
\end{split}
\end{equation}
where $\theta \in R^M$ are the learnable parameters and the $e_i \in R$ are zero mean, Gaussian, independent with variance of $\sigma^2$. 
The goal is to predict $y$ based on a new data sample $x$. 
Under the assumptions $y$, conditioned on $x$, has a normal distribution that depends on the learnable parameters $\theta$ 
\begin{equation}
p_{\theta}(y) 
=\frac{1}{\sqrt[]{2\pi\sigma^2}}\exp\left\{-\frac{1}{2\sigma^2}\big(y- x^T\theta \big)^2\right\}.
\end{equation}
The unknown parameter vector $\theta$ belongs to a set $\Theta$, which in the general case is the entire $R^M$. In the regularized version (leading to Ridge regression \cite{ridgeregression}), $\Theta$ is the sphere $|\theta|\leq A$. 
In the next section, the pNML will be evaluated for this hypotheses class. 
Recall that the pNML learner of $y$ given the the test sample $x$ and the training set $z^N=\{(x_i,y_i)\}_{i=1}^{N}$ is given by:
\begin{equation} \label{eq:pNML_def}
q_{\mbox{\tiny{pNML}}}(y|x;z^N) = \frac{1}{K} p_{\hat{\theta}(z^N;x,y)}(y|x).
\end{equation}
where in the linear regression case 
\begin{align}
\hat{\theta}(z^N,x,y)= \arg\min_{\theta\in\Theta} \left[ \sum_{i=1}^N \left(y_i - x_i^T\theta \right)^2 + \left(y-x^T\theta\right)^2 \right]
\end{align}
and where  $K$ is the the normalization factor:
\begin{equation} \label{gamma}
K = \int_R  p_{\hat{\theta}(z^N;x,y)}(y|x)dy,   
\end{equation}

The goal is to find an analytic expression for (\ref{eq:pNML_def}) and for the learnability measure $\Gamma = \log K$, the minmax regret value.

\section{pNML Evaluation} \label{sec:pNML_eval}
The following notation is used. 
$X \in R^{M \times N+1}$ is the matrix which contains all the training data along with the test sample and $Y \in R^{N+1}$ is the vector which contains all the labels including the test label, i.e.,
\begin{equation}
X = \begin{bmatrix} x_1 & \dots & x_N & x \end{bmatrix}, \;\;\ \ 
Y = \begin{bmatrix} y_1 \\ \vdots \\ y_N \\ y \end{bmatrix}
\end{equation}
Assuming that the test label $y$ is given, the optimal solution under least squares:
\begin{equation}
\hat{\theta}(z^N,x,y) = \theta^*_{N+1} = (X X^T)^{-1} X Y
\end{equation}
By the Recursive Least Squares (RLS) formulation \cite{hayes19969}:
\begin{equation} \label{eq:rls_update}
\theta ^*_{N+1} = \theta^*_{N} + P_{N+1} x (y - \hat{y})
\end{equation}
where $\hat{y} = x^T \theta ^*_{N}$ is the ERM prediction based on the training samples $\{(x_i, y_i)\}_{i=1}^{N}$ and\footnote{When $M > N$, $XX^T$ is not invertible, so $\lambda I$ with small $\lambda$ is added}
\begin{equation}
P_{N+1} = (XX^T)^{-1}. 
\end{equation}
Note that in RLS, $P_{N+1}$ is also calculated recursively from $P_N$, but this is not needed at this point.
Now,
\begin{equation}
\begin{split}
&p_{\theta_{N+1}^*}(y) 
=\frac{1}{\sqrt[]{2\pi\sigma^2}}\exp\left\{-\frac{1}{2\sigma^2}\big(y- x^T\theta_{N+1}^* \big)^2\right\} = \\
& \qquad \frac{1}{\sqrt[]{2\pi\sigma^2}}\exp\bigg\{-\frac{1}{2\sigma^2}\big(y - x^T \big(\theta^*_{N} + \\ 
& \qquad \qquad \qquad \qquad \qquad \qquad P_{N+1} x (y -\hat{y}) \big) \big)^2\bigg\} = \\
& \qquad \frac{1}{\sqrt[]{2\pi\sigma^2}}
\exp\left\{-\frac{(1 - x^T P_{N+1} x )^2 }{2\sigma^2}\left(y-\hat{y} \right)^2\right\}.  \\
\end{split}
\end{equation}
To get the pNML normalization factor (\ref{gamma}), we integrate over all possible labels
\begin{multline}
K = 
\int_{-\infty}^{\infty} \frac{1}{\sqrt[]{2\pi\sigma^2}}
\ exp\left\{-\frac{(1 - x^T P_{N+1} x )^2 }{2\sigma^2}
\left(y- \hat{y} \right)^2\right\} dy\\ 
=\frac{1}{1 - x^T P_{N+1} x } 
=\frac{1}{1 - x^T (XX^T)^{-1} x } \\
\end{multline}
Thus, the pNML distribution of $y$ given $x$ is:
\begin{multline}
q_{\mbox{\tiny{pNML}}}(y | x; z^N) = \frac{1}{K}p_{\theta_{N+1}^*}(y|x) = \\
\frac{1 - x^T P_{N+1} x }{\sqrt[]{2\pi\sigma^2}}
\exp\left\{-\frac{(1 - x^T P_{N+1} x )^2 }{2\sigma^2}\left(y-\hat{y} \right)^2\right\} \\
\end{multline}
and its associate learnability measure or regret:
\begin{equation} \label{eq:regret}
\Gamma = \log K = \log\left(\frac{1}{1 - x^T (XX^T)^{-1} x } \right).
\end{equation}

\section{pNML with Regularization} \label{sec:pNMLwithReg}
Next, we shall assume that the model class $\Theta$ is constrained to the sphere  $|\theta|\leq A$, for some $A$. 
Using a Lagrange multiplier $\lambda$ we get the Tikhonov regularization (or Ridge regression), where the expression to minimize is now: 
\begin{equation}
\mathcal{L}(z^N)= \sum_{i=1}^{N}|y_i-x_i^T \theta|^2 + \lambda |\theta|^2
\end{equation}
With the test data, the ``regularized'' least square solution is:
\begin{equation}
\hat{\theta}(z^N,x,y) = \theta_{N+1}^* = (X X^T+ \lambda I)^{-1} X Y
\end{equation}
Here too the RLS formula holds: 
\begin{equation}
\theta_{N+1}^*=\theta^*_{N} + P_{N+1} x (y - \hat{y})
\end{equation}
However, now 
\begin{equation}
P_{N+1}= (X X^T+ \lambda I)^{-1}.    
\end{equation}
The rest of the evaulation is similar to \ref{sec:pNML_eval}, yielding the following pNML learner:
\begin{multline} \label{eq:pNML_least_sqaures}
q_{\mbox{\tiny{pNML}}}(y|x; z^N, \lambda)
=\frac{1 - x^T (XX^T + \lambda I)^{-1} x }{\sqrt[]{2\pi\sigma^2}} \\
\cdot \exp\left\{-\frac{(1 - x^T (XX^T + \lambda I)^{-1} x )^2 }{2\sigma^2}\left(y- \hat{y} \right)^2\right\} \\
\end{multline}
and the associated regret or the log-normalization factor:
\begin{equation}
\Gamma = \log K = \log \left( \frac{1}{1 - x^T (XX^T + \lambda I)^{-1} x } \right)
\end{equation}
Note that regularization can help in the case where $XX^T$, the unnormalized correlation matrix of the data is
ill conditioned. 
In the next section we find the ``learnable space'' for the linear regression problem and observe situations where this regularization is needed.

\section{Learnable Space} \label{sec:learnable_space}
In order to understand for which test sample the trained model generalizes well we need to look at the regret expression (\ref{eq:regret}). 
High regret means that the pNML learner is very far from the genie and therefore we may not trust its predictions. 
Low regret, on the other hand, means the model is as good as a genie who knows the true test label, and so it is trusted.

Consider the matrix $X_N = [x_1,x_2, \hdots, x_{N}]$, composed of the training data, and apply the singular value decomposition (SVD) on it, i.e., $X_N = U \Sigma V^T$ with $U\in R^{M \times M}$, 
$\Sigma$ is a rectangular diagonal matrix of the singular values and $V \in R^{N \times N}$. 
The expression $x^T(XX^T)^{-1}x$ can be rewritten as:
\begin{multline}
x^T(XX^T)^{-1}x = 
x^T\left(\begin{bmatrix} U \Sigma V^T & x \end{bmatrix}
\begin{bmatrix}
V \Sigma^T U^T \\ x^T
\end{bmatrix}
\right)^{-1}x \\
=  x^T\left(U \Sigma \Sigma^T U^T + x x^T\right)^{-1}x.
\end{multline}
Denote by $R_N$ the empirical correlation matrix of the training: 
\begin{equation}
R_N=\frac{1}{N} U \Sigma \Sigma^T U^T=U H U^T \ \ R_N^{-1}=  U H^{-1} U^T   
\end{equation}
where H is a diagonal matrix with $H_{ii}=\eta_i$, the eigenvalues of $R_N$.
By the matrix inversion lemma, see \cite{press2007section}, 
we have:
\begin{equation}
x^T(XX^T)^{-1}x = 
x^T \left[ \frac{1}{N}R_N^{-1} -  \frac{\frac{1}{N^2}R_N^{-1} x x^T  R_N^{-1}}{1 + \frac{1}{N} x^T  R_N^{-1} x} \right] x.
\end{equation}
Denote $\gamma = x^T R_N^{-1} x$. We can simplify the expression:
\begin{equation}
x^T(XX^T)^{-1}x = \frac{1}{N}\gamma - \frac{\frac{1}{N^2}\gamma^2}{1+\frac{1}{N}\gamma} = \frac{\frac{1}{N}\gamma}{1+\frac{1}{N}\gamma}.
\end{equation}
Plugging in the regret formula (\ref{eq:regret}):
\begin{equation}
\Gamma = \log K = \log \left( \frac{1}{1-\frac{\frac{1}{N}\gamma}{1+\frac{1}{N}\gamma}} \right)
=  \log \left( 1+\frac{1}{N} \gamma \right).
\end{equation}
Let $u_i$ be the eigenvectors of the empirical correlation matrix of the training data. 
Expressing $\gamma$ by $x^Tu_i$, the projections of $x$ on $u_i$:
\begin{multline}
\gamma = 
\begin{bmatrix}
x^T u_1 & \hdots & x^T u_M
\end{bmatrix}
\begin{bmatrix}
\frac{1}{\eta_1} & \hdots & 0 \\
\vdots & \vdots &  \vdots \\
0 & \hdots &  \frac{1}{\eta_M} \\
\end{bmatrix}
\begin{bmatrix}
u_1^T x \\ \vdots \\ u_M^T x
\end{bmatrix} \\
= \sum_{i=1}^{M} \frac{\left(x^Tu_i\right)^2}{\eta_i}.
\end{multline}
The final regret expression is thus:
\begin{equation}
\Gamma = \log K = \log \left(1 + \frac{1}{N} \sum_{i=0}^{M} \frac{\left(x^Tu_i\right)^2 }{\eta_i}\right).
\end{equation}
If the test sample $x$ lies mostly in the subspace spanned by the eigenvectors with large eigenvalues, then the model can generalize well even if $M>N$.

\section{Simulation} \label{sec:simulation}
In this section we present some simulations that demonstrate the results above. 
We chose the problem of fitting a polynomial to data, which is a special case of linear regression.
The simulations show prediction and generalization capabilities in a variety of regularization factors and polynomial degrees.

\begin{figure}[tb] 
    \centering
    \includegraphics[width=\linewidth]{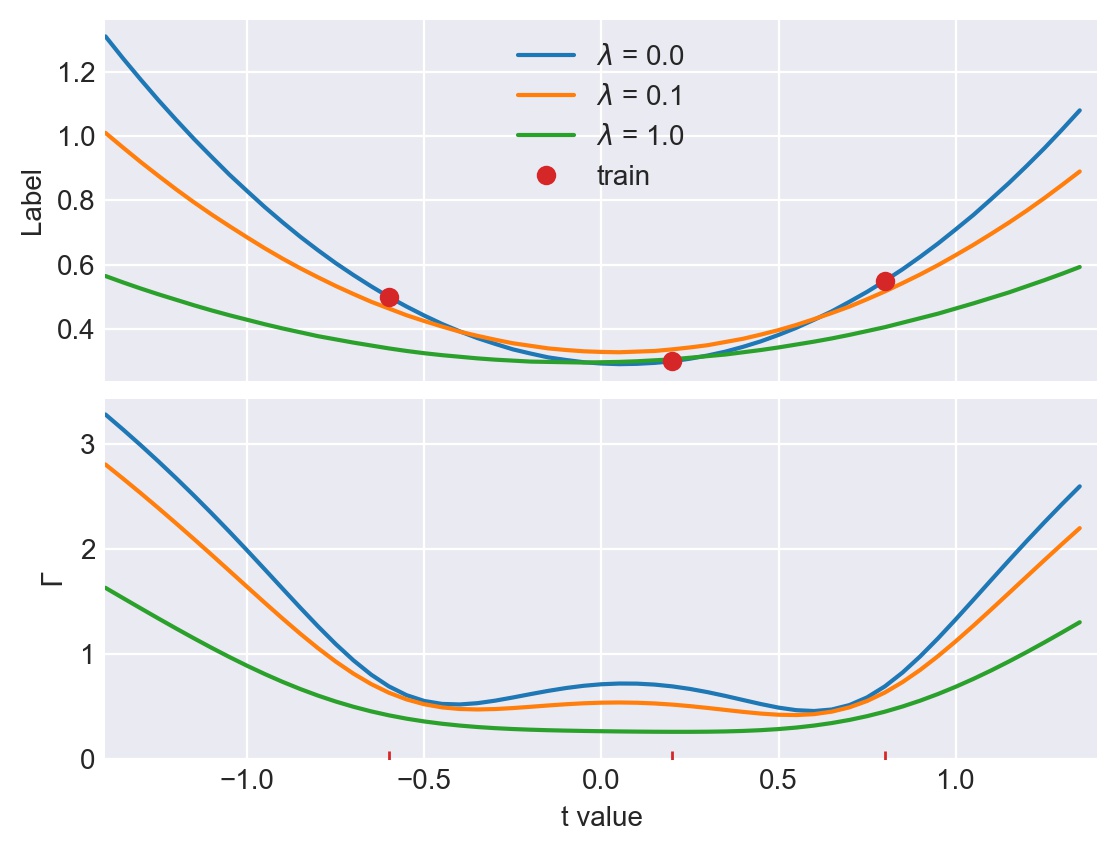}
    \caption{\textbf{Least squares predictor with variety of regularization terms.} (Top) The least squares estimator fitted to the training data (in red) with different values of regularization term. (Bottom) The regret of the pNML learner from (\ref{eq:regret}) on the interval [-1,1]. The training data are marked in red on the x axis.}
    \label{fig:least_squares_with_reg}
\end{figure}

In the first experiment we generated 3 random points, $t_0,t_1,t_2$, uniformly in the interval $[-1, 1]$. These points are the training set and are shown in Figure \ref{fig:least_squares_with_reg} (top) as red dots. 
The relation between $y$ and $t$ is given by a polynomial of degree two.
Thus, the X matrix of section \ref{sec:formal_problem_def} is given by:
\begin{equation} \label{eq:two_degree_pol}
X = 
\begin{bmatrix}
1 & 1 & 1 \\
t_0 & t_1 & t_2 \\
t_0^2 & t_1^2 & t_2^2 
\end{bmatrix}.
\end{equation}
Based on the training we predict a probability for all t values in the interval [-1,1] using (\ref{eq:pNML_least_sqaures}) with a regularization factor $\lambda$ of $0$, $0.1$ and $1.0$. 
It is shown in Figure \ref{fig:least_squares_with_reg} (top) that without regularization ($\lambda=0$), the blue curve fits the data exactly. As $\lambda$ increases the fitted curve becomes less steep but tends to fit less to the training data.

Figure \ref{fig:least_squares_with_reg} (bottom) shows the regret, given by (\ref{eq:regret}), for the polynomial model from \eqref{eq:two_degree_pol} for all $t\in[-1,1]$ where the training $t_i$'s are marked in red on the x axis. 
We can see that around the training data the regret is very low in comparison to areas where there are no training data. 
In addition, models with larger regularization term have lower regret for every point in the interval $[-1,1]$.
For all regularization terms, the regret increases as moving away from the training data.

\begin{figure}[tb]
    \centering
    \includegraphics[width=\linewidth]{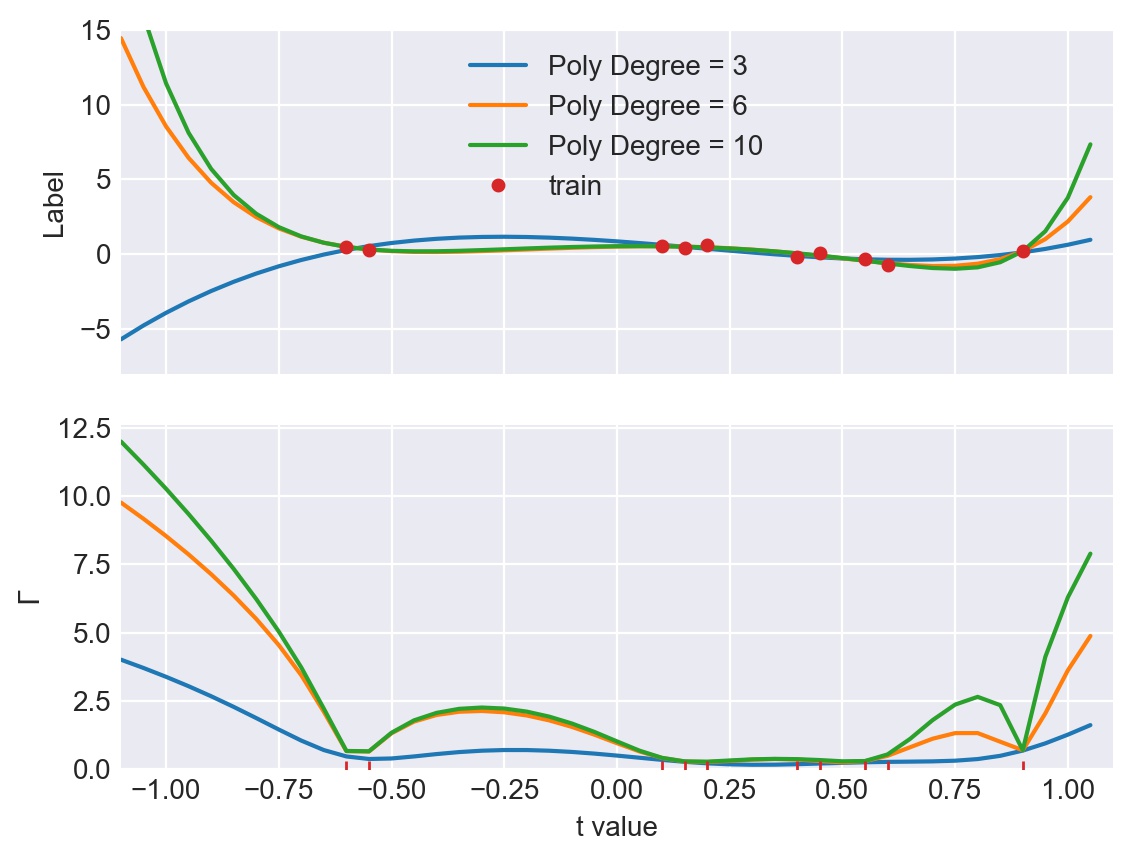}
    \caption{\textbf{Least squares predictor with different polynomial degree.} (Top) pNML least squares predictions with different polynomial degrees. (Bottom) The regret of the pNML learners from (\ref{eq:regret}) on the interval [-1,1]. The training data t values are marked in red on the x axis.}
    \label{fig:least_squares_with_poly}
\end{figure}

Next, we simulate the case of fitting polynomials with different degrees. 
Again, we generated 10 random points in the interval $[-1, 1]$.
The matrix $X$ is now:
\begin{equation}
X = 
\begin{bmatrix}
1 & 1 & 1 & \hdots & 1\\
t_0 & t_1 & t_2 & \hdots & t_{9}\\
\vdots &  \vdots &       & \vdots  \\
t_0^{\textit{Poly Deg}} & t_1^{\textit{Poly Deg}} & t_2^{\textit{Poly Deg}}  & \hdots & t_{9}^{\textit{Poly Deg}}
\end{bmatrix}.
\end{equation}
Figure \ref{fig:least_squares_with_poly} (top) shows the predicted label for every $t$ value in $[-1,1]$ for the different polynomial degrees. 
To avoid singularities we used the regularized version with $\lambda=10^{-4}$. 
The training set is shown by red dots in the figure. 
Note that for a polynomial of degree ten, the number of parameters is greater than the size of the training set. 
Nevertheless, the prediction accuracy near the training samples is similar to that of a degree three polynomial.
Figure \ref{fig:least_squares_with_poly} (bottom) shows the regret (or learnability) of the three pNML learners corresponding to model classes of polynomials with the various degrees. 
All the learners have regret values that are small near the training samples and large as $t$ drifts away from these samples.

\section{Conclusions} \label{sec:conclusion}

In this paper, we provided an explicit analytical solution of the pNML universal learning scheme and its learnability measure for the linear regression hypothesis class. 
Interestingly, the predicted universal pNML assignment is Gaussian with a mean that is equal to that of the ERM, but with a variance that increases by a factor $K$ whose logarithm is the learnability measure $\Gamma$.
Analyzing $\Gamma$ we can observe the ``learnability space'' for this problem. 
Specifically, if a test sample mostly lies in the subspace spanned by the eigenvectors associated with large eigenvalues of the empirical correlation matrix then the learner can generalize well, even in an over-parameterized case where the regression dimension is larger than the number of training samples.
Finally, we provided a simulation of the pNML least squares prediction for polynomial interpolation.

This work suggests a number of potential directions for future work, some are already explored in an accompanying paper \cite{pNML_neural_networks}. 
We conjecture that as in linear regression other ``over-parameterized'' model classes are learnable at least locally, that can be inferred from the pNML solution. 
This notion is indeed corroborated by the findings in \cite{pNML_neural_networks}.



\bibliographystyle{IEEEtran}
\bibliography{main_pnml_linear_regression}

\begin{thebibliography}{10}
\providecommand{\url}[1]{#1}
\csname url@samestyle\endcsname
\providecommand{\newblock}{\relax}
\providecommand{\bibinfo}[2]{#2}
\providecommand{\BIBentrySTDinterwordspacing}{\spaceskip=0pt\relax}
\providecommand{\BIBentryALTinterwordstretchfactor}{4}
\providecommand{\BIBentryALTinterwordspacing}{\spaceskip=\fontdimen2\font plus
\BIBentryALTinterwordstretchfactor\fontdimen3\font minus
  \fontdimen4\font\relax}
\providecommand{\BIBforeignlanguage}[2]{{%
\expandafter\ifx\csname l@#1\endcsname\relax
\typeout{** WARNING: IEEEtran.bst: No hyphenation pattern has been}%
\typeout{** loaded for the language `#1'. Using the pattern for}%
\typeout{** the default language instead.}%
\else
\language=\csname l@#1\endcsname
\fi
#2}}
\providecommand{\BIBdecl}{\relax}
\BIBdecl

\bibitem{lawson1995solving}
C.~L. Lawson and R.~J. Hanson, \emph{Solving least squares problems}.\hskip 1em
  plus 0.5em minus 0.4em\relax Siam, 1995, vol.~15.

\bibitem{james2013introduction}
G.~James, D.~Witten, T.~Hastie, and R.~Tibshirani, \emph{An introduction to
  statistical learning}.\hskip 1em plus 0.5em minus 0.4em\relax Springer, 2013,
  vol. 112.

\bibitem{universal_prediction}
N.~Merhav and M.~Feder, ``Universal prediction,'' \emph{IEEE Transactions on
  Information Theory}, vol.~44, no.~6, pp. 2124--2147, 1998.

\bibitem{FogelFeder2018}
Y.~Fogel and M.~Feder, ``Universal supervised learning for individual data,''
  \emph{arXiv preprint arXiv:1812.09520}, 2018.

\bibitem{valiant1984theory}
L.~G. Valiant, ``A theory of the learnable,'' \emph{Communications of the ACM},
  vol.~27, no.~11, pp. 1134--1142, 1984.

\bibitem{shtar1987universal}
Y.~M. Shtar'kov, ``Universal sequential coding of single messages,''
  \emph{Problemy Peredachi Informatsii}, vol.~23, no.~3, pp. 3--17, 1987.

\bibitem{roos2008sequentially}
T.~Roos and J.~Rissanen, ``On sequentially normalized maximum likelihood
  models,'' 2008.

\bibitem{roos2008bayesian}
T.~Roos, T.~Silander, P.~Kontkanen, and P.~Myllymaki, ``Bayesian network
  structure learning using factorized nml universal models,'' in
  \emph{Information Theory and Applications Workshop, 2008}.\hskip 1em plus
  0.5em minus 0.4em\relax IEEE, 2008, pp. 272--276.

\bibitem{vapnik1992principles}
V.~Vapnik, ``Principles of risk minimization for learning theory,'' in
  \emph{Advances in neural information processing systems}, 1992, pp. 831--838.

\bibitem{vapnik2013nature}
------, \emph{The nature of statistical learning theory}.\hskip 1em plus 0.5em
  minus 0.4em\relax Springer science \& business media, 2013.

\bibitem{sontag1998vc}
E.~D. Sontag, ``Vc dimension of neural networks,'' \emph{NATO ASI Series F
  Computer and Systems Sciences}, vol. 168, pp. 69--96, 1998.

\bibitem{hayes19969}
M.~H. Hayes, ``9.4: Recursive least squares,'' \emph{Statistical Digital Signal
  Processing and Modeling}, p. 541, 1996.

\bibitem{cardinali2013observation}
C.~Cardinali, ``Observation influence diagnostic of a data assimilation
  system,'' in \emph{Data Assimilation for Atmospheric, Oceanic and Hydrologic
  Applications (Vol. II)}.\hskip 1em plus 0.5em minus 0.4em\relax Springer,
  2013, pp. 89--110.

\bibitem{trevor2009elements}
H.~Trevor, T.~Robert, and F.~JH, ``The elements of statistical learning: data
  mining, inference, and prediction,'' 2009.

\bibitem{ridgeregression}
A.~Hoerl and R.~Kennard, ``Ridge regression: Biased estimation for
  nonorthogonal problems,'' \emph{Technometrics}, vol.~12, 1970.

\bibitem{press2007section}
W.~H. Press, S.~A. Teukolsky, W.~T. Vetterling, and B.~P. Flannery, ``Section
  2.7. 1 sherman--morrison formula,'' \emph{Numerical Recipes: The Art of
  Scientific Computing (3rd ed.). Cambridge University Press, New York},
  vol.~1, pp. 55--67, 2007.

\bibitem{pNML_neural_networks}
K.~Bibas, Y.~Fogel, and M.~Feder, ``Deep pnml: Predictive normalized maximum
  likelihood for deep neural networks,'' 2019.

\end{thebibliography}

\end{document}